# An Empirical Evaluation of a Randomized Algorithm for Probabilistic Inference

R. Martin Chavez and Gregory F. Cooper
Section on Medical Informatics
Stanford University School of Medicine
Stanford, California 94305

## Abstract

In recent years, researchers in decision analysis and artificial intelligence (AI) have used Bayesian belief networks to build models of expert opinion. Using standard methods drawn from the theory of computational complexity, workers in the field have shown that the problem of probabilistic inference in belief networks is difficult and almost certainly intractable. KNET, a software environment for constructing knowledge-based systems within the axiomatic framework of decision theory, contains a randomized approximation scheme for probabilistic inference. The algorithm can, in many circumstances, perform efficient approximate inference in large and richly interconnected models of medical diagnosis. Unlike previously described stochastic algorithms for probabilistic inference, the randomized approximation scheme computes a priori bounds on running time by analyzing the structure and contents of the belief network.

In this article, we describe a randomized algorithm for probabilistic inference and analyze its performance mathematically. Then, we devote the major portion of the paper to a discussion of the algorithm's empirical behavior. The results indicate that the generation of good trials (that is, trials whose distribution closely matches the true distribution), rather than the computation of numerous mediocre trials, dominates the performance of stochastic simulation.

## 1. Introduction

$\mathcal{P}$ denotes the set of all decision problems that a *deterministic* Turing machine can answer in a period of time bounded by a polynomial in the size of the problem description. $\mathcal{NP}$ is the set of all decision problems that a *nondeterministic* Turing machine can resolve in polynomial time. Deterministic machines pursue one path of computation at a time; nondeterministic machines can perform multiple computations simultaneously.

The hardest problems in $\mathcal{NP}$, known as $\mathcal{NP}$-complete, probably do not admit polynomial-time deterministic algorithms (Garey and Johnson, 1979). If any of the $\mathcal{NP}$-complete problems admits a polynomial-time deterministic algorithm, then $\mathcal{P} = \mathcal{NP}$, and all problems in $\mathcal{NP}$ can be solved in polynomial time. The vast majority of theoreticians believe, however, that $\mathcal{NP}$ properly contains $\mathcal{P}$, and that polynomial-time algorithms for $\mathcal{NP}$-complete problems do not exist. $\mathcal{NP}$-hard problems are at least as difficult to answer as are the $\mathcal{NP}$-complete problems, if not more so.

Given truth assignments for a set $E$ of random variables in a belief network, an algorithm for Probabilistic Inference in Belief NETworks (PIBNET) computes the posterior probabilities for the outcomes of a specified node $X$. PIBNET is hard for $\mathcal{NP}$, by reduction from 3-satisfiability in the propositional calculus (Cooper, 1987). The classification of PIBNET as $\mathcal{NP}$-hard has prompted a shift in focus away from deterministic algorithms and toward approximate methods, heuristics, and analyses of average-case behavior.

There now exists a number of algorithms for probabilistic inference in belief networks: the message-passing algorithm of Pearl (Pearl, 1986), the triangulation method of Lauritzen and Spiegelhalter (Lauritzen and Spiegelhalter, 1987), and the randomized approximation scheme (ras) described herein. Each algorithm has computational properties that render it attractive for inference on certain kinds of networks. The $\mathcal{NP}$-hard classification suggests, however, that no algorithm can provide a definitive efficient solution for all inference problems.

The ras, in particular, builds upon straight stochastic simulation as proposed by Pearl and others (Pearl, 1987b; Pearl, 1987a). Stochastic simulations generate *trials*, or instantiations of random variables governed by a joint probability distribution, and then use the frequencies of random outcomes to approximate posterior probabilities. Two kinds of error can plague such simulations: The distribution of generated trials does not necessarily correspond closely to the true distribution of outcomes, and a paucity of trials can give rise to sampling error. We must focus not only on the generation of myriad trials, but also on the quality of those trials.

Unlike previous methods of straight simulation, our randomized algorithm (known hereafter as BN-RAS) gives precise a priori bounds for its running time as a function of relative or interval error.[1] The ras does not necessarily run faster than previous simulation algorithms. Indeed, the algorithm itself proposes only minor modifications to straight simulation. Those modifications, however, permit a theoretical convergence analysis that does not apply to the original asymmetric methods of straight simulation.

[1]In the literature of theoretical computer science, our ras would be classified as a fully polynomial randomized approximation scheme (fpras). Although the term is a standard one in the study of algorithms, it may mislead those less familiar with the nomenclature. We emphasize that our ras exhibits performance that varies linearly with the topological complexity of the network, even though the same algorithm degrades dramatically as the probabilities themselves approach 0 and 1.

By definition, an ras computes approximate answers that, with *correctness probability* greater than $1-\delta$, differ from the true answer by a *relative error* of no more than $\epsilon$ (Karp and Luby, 1983). In addition, an ras requires computing time that is a polynomial in $1/\epsilon$, $1/\delta$, and the size of the input.[2] Modified versions of the ras guarantee, with high probability $1-\delta$, upper bounds on *interval error* $\alpha$, to be distinguised from the relative error $\epsilon$.[3]

In this paper, we briefly summarize the salient characteristics of BN-RAS and present an empirical investigation of its properties. In particular, we focus attention on the computational complexity of generating good trials.

# 2. Methods and Procedures

## 2.1 Approximate Probabilistic Inference

We now offer a complexity-theoretic treatment of approximate probabilistic inference. We use methods drawn from the analysis of ergodic Markov chains and randomized complexity theory to build an algorithm that approximates the solutions of inference problems for many belief networks to within arbitrary precision. We slightly alter a previous simulation scheme designed by Pearl (Pearl, 1987b; Pearl,

[2]In the case of belief networks, the size of the input is the length of a string that fully describes the nodes, their connections, and their conditional probabilities, represented in unary notation. Unary notation ensures a running time that is a polynomial in the problem size; as the probabilities approach 0 and 1, the unary representations assume unbounded size, and the algorithm's performance decreases dramatically. Note, in addition, that we have excluded deterministic relationships (probabilities equal to 0 or 1) from the analysis.

[3]The interval error $\alpha$ is the maximum difference between the true and the approximate probabilities, taken over all the probabilities in the network. The relative error $\epsilon$ is the maximum difference between the true and the approximate probabilities, divided by the true probability. In general, it is much more difficult to guarantee a fixed relative error, especially as the probabilities approach 0.

1987a), known hereafter as the method of *straight simulation*, so as to render an analysis of its computational characteristics. We have given the full derivation in (Chavez, 1989b; Chavez, 1989a), and present only the salient results here.

Suppose that we wish to compute all posterior probabilities in the network to within an interval error $\alpha$. Suppose, in addition, that we are willing to tolerate a small probability $\delta$ that the algorithm fails to converge within the $\alpha$ bound. The detailed argument, based on Chebyshev's inequality and the scheme of Karp and Luby, reveals that

$$N \geq 1/(4\delta\alpha^2) \tag{1}$$

guarantees the $(\alpha, \delta)$ convergence criteria, where $N$ is the total number of trials. Each trial corresponds to the choice of a joint instantation for all the nodes in the belief network.

We have predicated our analysis on the existence of a trial generator that accurately produces states of the network according to their true probabilities, contingent on the available evidence. The original straight-simulation generator depends on the initial state (that is, it lacks ergodicity). Moreover, the straight-simulation generator offers no guarantees about its convergence properties. We must, therefore, turn our attention to the study of modified state generators.

Given any belief network, we show how to construct a special Markov chain with the following two properties. First, states of the Markov chain correspond to joint instantiations of all the nodes in the network; the Markov chain associated with a network of $n$ binary nodes, for example, has $2^n$ distinct states. Second, the stationary distribution of the Markov chain is identical to the joint posterior-probability distribution of the underlying belief network.

In addition, the constructed Markov chain has the properties of *ergodicity* and *time reversibility*. Ergodic chains are, by definition, *aperiodic* (without cycles) and *irreducible* (with a nonzero transition probability between any pair of states). Time-reversible chains look the same whether the simulation flows forward or backward. Once again, (Chavez, 1989a) presents the details of the construction.

In the limit of infinity, after the Markov chain has reached its stationary distribution, that chain generates states according to their true probabilities. Obviously, we cannot afford to let the chain reach equilibrium at infinity. In practice, we wish to know how well the chain has converged after we have let it run for only a finite number $t$ of transitions. Define the *relative pointwise distance* (r.p.d.) after $t$ transitions,

$$\Delta(t) = \max_{i,j \in [M]} \left\{ \frac{\left| P_{ij}^{(t)} - \pi_j \right|}{\pi_j} \right\},$$

where $P_{ij}^{(t)}$ denotes the $t$-step transition probability from state $i$ to $j$ (with a total of $M$ states) and $\pi_j$ denotes the stationary probability of state $j$. Let $\Pi = \min_{i \in [M]} \pi_i$, the joint probability of the least likely joint state of all variables in the network. We wish to determine the number of transitions $t$ needed to guarantee a deterministic upper bound on $\Delta(t)$. A Jerrum-Sinclair analysis of chain conductance (intuitively, the chain's tendency to flow around the state space (Jerrum and Sinclair, 1988)) and a combinatoric path-counting argument show that the BN-RAS generator requires

$$t \geq \frac{\log \gamma + \log \Pi}{\log(1 - p_0^2/8)} \tag{2}$$

transitions to guarantee a relative pointwise distance of $\gamma$, where $p_0$ is the smallest transition probability in the Markov chain.

Combining the convergence analysis with the scoring strategy in relation (1), BN-RAS computes posterior-probability estimators $\tilde{Y}$ that satisfy the constraint

$$\frac{\Pr[e|\xi]}{(1+\gamma)} - \alpha \leq \tilde{Y} \leq (1+\gamma)\Pr[e|\xi] + \alpha \tag{3}$$

with probability greater than $1 - \delta$. To do so, the algorithm must perform $t$ transitions per trial, with

$$t = \left\lceil \frac{4(1+\gamma)^3}{3\alpha^2} \right\rceil \cdot (12\lceil -\log\delta \rceil + 1) \cdot \frac{\log \gamma + \log \Pi}{\log(1 - p_0^2/8)}, \tag{4}$$

where each transition corresponds to one step of the underlying Markov chain. We can then use those posterior-probability estimators to rank the leading

```
/* Compute a transition p.m.f. */
void sample(this)
node *this;
{
    for (sum = 0.0,
         this->value = 0;
         this->value < this->nvalues; v++)
    {
        /* compute P[this | parents] */
        prod = cprob(this);
        /* multiply by P[child | this],
           for each child */
        for (i = 0; i < this->nchildren; i++)
           prod *=
              cprob(matrix[this->children[i]]);
        sum += prod;
        this->dist[v] = sum;
    }
    normalize dist;
}
```

Figure 1: Pseudo-code that computes the Markov transition probabilities from a belief network.

diagnoses. Thus, BN-RAS efficiently computes approximate inferences within the normative framework of probability theory, so long as $p_0$ is not too small.

Conceptually, now, the ras has a simple description. Each of $N$ *trials* produces a joint instantiation of nodes in the belief network. To conduct each trial, we initialize all nodes to random values from a uniform distribution and we run the chain for $t$ *transitions*. [4] We compute each transition probability with the pseudo-code given in Figure 1. The $t$ transitions per trial help to ensure mixing of the underlying Markov chain, and thereby facilitate the generation of good trials (that is, trials from a distribution that approximates the true distribution of states to within a bounded relative error). The theoretical analysis, in short, gives worst-case formulae for $N$ and $t$ as a function of network parameters and error tolerance.

[4]The random initialization is not required for the analysis, but such an initialization can occasionally improve the performance of the simulation.

We present a pseudo-code description of BN-RAS in Figure 2. In order to obtain the original straight simulation, we replace the routines `do_transition()` and `next_trial()` with the variants in Figure 3. Specifically, BN-RAS exhibits the following differences from straight simulation: (1) BN-RAS chooses a random state in the Markov chain before computing each trial; (2) BN-RAS makes transitions in a manner that renders the Markov chain aperiodic, irreducible, and time-reversible; (3) BN-RAS makes many transitions before scoring a single trial, whereas straight simulation scores the result of every transition. Those properties allow us to analyze the algorithmic complexity of BN-RAS in great detail. No general techniques exist, however, for analyzing time-irreversible and non-ergodic Markov chains such as the chains used in straight simulation.

The preceeding analysis of complexity clarifies the underlying computational properties of the algorithm, but it says little about the method's performance on examples drawn from the real world. We now describe the complexity analysis in greater detail, and answer questions about the algorithm's performance on two examples. We use the exact algorithm of Lauritzen and Spiegelhalter (Lauritzen and Spiegelhalter, 1987) to compute the gold standard for our comparisons.

With the analysis of BN-RAS in hand, we propose to address the following questions:

- How does the error change as the number of transitions $t$ per trial increases? (We have an interest in two error measures: the average error over all nodes, and the interval error for the node with the greatest discrepancy from the gold standard.)

- How does the error change as the number of trials increases, with the number of transitions $t$ per trial held constant?

```
/* Perform one transition */
do_transition()
{
    /* With probability 1/2,
       stay put (guarantees
       aperiodicity).
       Use ACM algorithm. */
    p = drand_acm();
    if (p <= 0.5)
      return;
    else
    {
      uniformly choose a node 'this';
       /* Compute the transition p.m.f.,
          and choose a value */
      sample(this);
      this->value = choose(this->dist);
    }
}

/* Compute a trial */
next_trial()
{
    set all the nodes to uniform
       random values;
    for (d = 0; d < transitions; d++)
      do_transition();
}

estimate()
{
    compute number of trials, n;
    for (j = 1; j <= n; j++)
    {
      nextstate();
       score outcomes;
    }
}
```

Figure 2: The pseudo-code for BN-RAS.

```
/* Compute a transition probability, and
   assign a value to node i */
do_transition(i)
{
    sample(matrix[i]);
    matrix[i]->value =
       choose(matrix[i]->dist);
}

/* Compute a trial */
next_trial()
{
    static int i = 0;

    /* Don't re-initialize
       the nodes */
    if (i == nnodes)
       i = 0;
    do_transition(i++);
}
```

Figure 3: The pseudo-code for straight simulation.

- We predicated the analysis of BN-RAS on a set of distribution-free, worst-case assumptions. Does the computation time required for reasonable convergence in specific cases undercut the analytic estimates?
- BN-RAS throws away many of its generated trials. Does the sampling of fewer trials reduce the efficiency of the approach? In other words, how does BN-RAS perform in comparison to straight simulation?

We study each of those questions in turn, and display charts and graphs that illustrate our conclusions.

## 3. Results

In the present experiments, we study two belief networks: a simple two-node network (Figure 4) for which straight simulation is known to perform poorly (Chin and Cooper, 1987), and a much more complex network, DxNet, for alarm management in the intensive-care unit (Beinlich *et al.*, 1988).

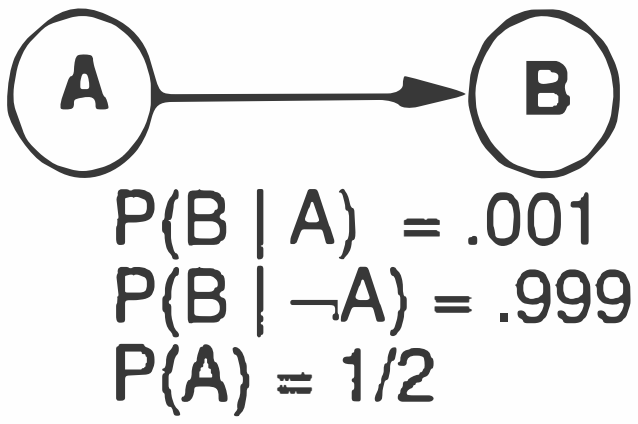

Figure 4: This two-node network poses severe convergence problems for straight simulation.

The two-node network specifically causes straight simulation to undergo pathological oscillation. DxNET, on the other hand, reflects an anesthesiologist's clinical expertise and judgmental knowledge. For our present purposes, we observe that, to guarantee $\epsilon = 0.1$, $\gamma = 0.1$, and $\delta = 0.1$ for the two-node example, we require 2,662,000 trials, with 316,911,596 transitions per trial; to guarantee an interval error $\alpha = 0.1$ for the same network, we need only 1332 trials, with the same number of transitions per trial. For DxNET, the numbers prove even more formidable. The worst-case bounds require 13,307,782 trials, with 256,573,353,901 transitions per trial, to guarantee $\epsilon = 0.1$, $\gamma = 0.1$, and $\delta = 0.1$; to ensure that $\alpha = 0.1$, we need only 1332 trials, but we still require 256,573,353,901 transitions per trial.

Figure 5 illustrates the number of trials, based on relation (1), as a function of the interval error $\alpha$ for several values of the failure probability $\delta$. With an error tolerance of $\alpha = 0.1$, the algorithm requires less than 10,000 trials, for values of $\delta > 0.1$. As the error tolerance shrinks, however, the number of trials increases quadratically. Note, however, that relation (1) specifies a distribution-free upper bound on the number of trials. Depending on the underlying probability distribution, fewer trials may suffice.

Figure 6 illustrates the relationship between the number of transitions needed for sufficient mixing of the Markov chain, $t$, and the smallest transition probability, $p_0$. The transition probabilities vary as the product of conditional probabilities at each local node group. The belief networks that knowledge engineers build for realistic applications will typically require small transition probabilities. Such probabilities do not entail the approximation scheme's success or failure; rather, they suggest that the analytic bounds cannot guarantee efficient computation. Note, in particular, the logarithmic abscissa, and the relative unimportance of the $\gamma$ error term. For belief networks in which the smallest transition probability $p_0 \geq 0.1$, we expect that BN-RAS will yield an acceptable, tractable computation for realistic values of $\alpha$ and $\delta$. As $p_0$ approaches 0, however, the number of transitions needed to guarantee the bounds, as indicated in equation (4), approaches infinity.

Clearly, the analytic bounds do not always yield an efficient algorithm, even though they do predict a running time that varies only linearly with the number of nodes. The conditional probabilities that lie close to 0 and 1 require unrealistically large values of $t$ to approximate the stationary distribution with great certainty.

## 3.1 DxNet Performance Measurements and Time Complexity

In this section, we study the performance of BN-RAS for the DxNET problem on a Sun-4 timesharing processor running SunOS, a version of 4.3bsd UNIX. We measured CPU time with the UNIX system call `clock()`, which returns the elasped processor time in microseconds.

Figures 7 and 8 demonstrate that the CPU time increases linearly with the number of trials and the number of transitions per trial, as expected. Those figures serve as nomograms for translating $N$ and $t$ into realistic CPU-consumption figures on the Sun-4.

Figures 9 illustrates the most crucial insight of this empirical study. The convergence depends not so much on the number of tabulated trials, but rather on the quality of those trials (as determined by the number of transitions per trial). In other words, if we had an ideal trial generator, we could expect very rapid convergence; inasmuch as the raw Markov chain reaches the stationary distribution only after many thousands of transitions, however, trial generation in DxNET poses the greatest difficulty. If we could somehow modify the Markov chain and thereby increase the rate at which it reaches the stationary distribution, or if we could compute an ini-tial state from which the chain converges to the limit in just a few transitions, the rate of convergence of BN-RAS would greatly increase.

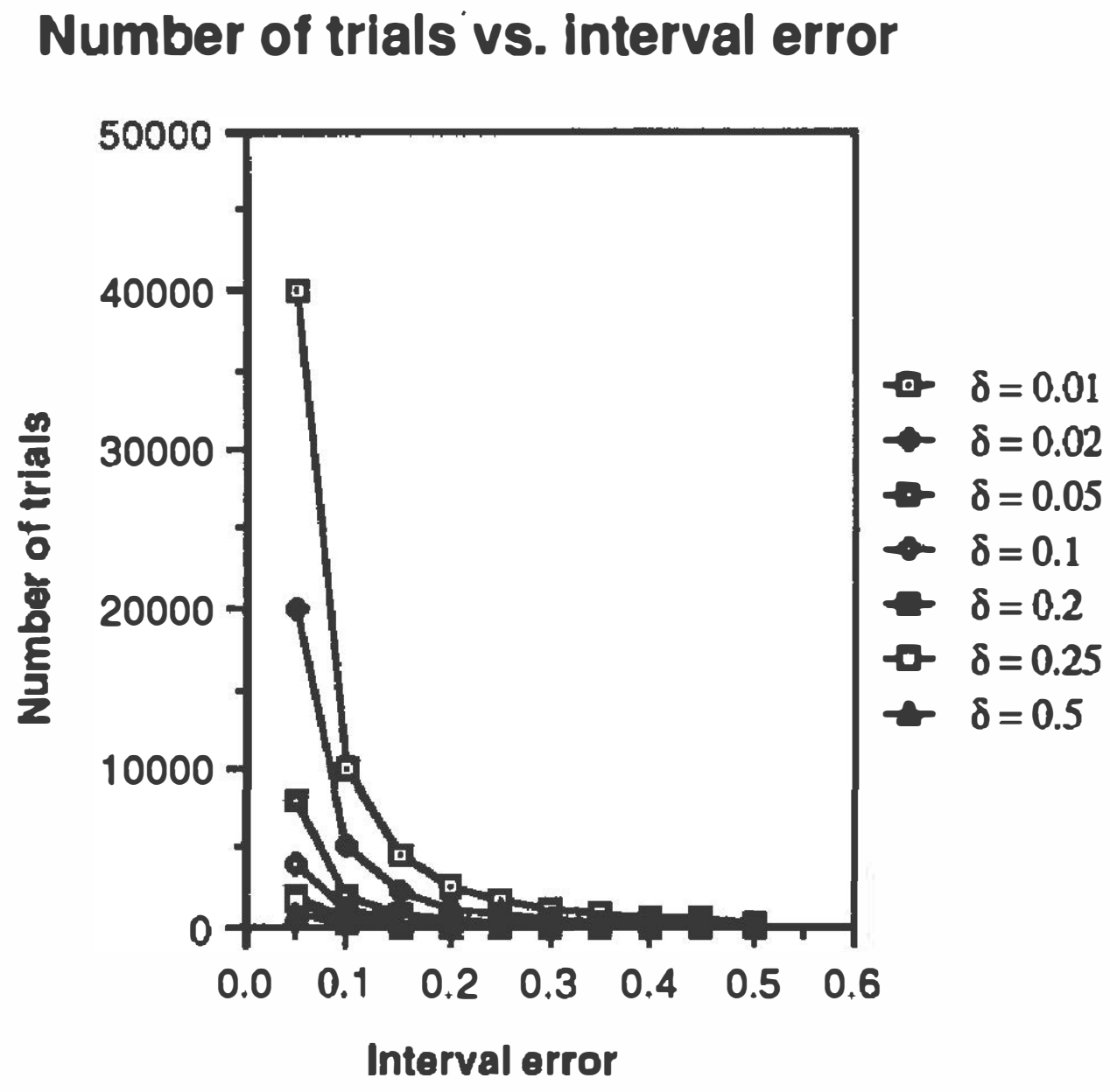

Figure 5: This graph demonstrates the relationship between the number of trials $N$ needed to guarantee an $(\alpha, \delta)$ algorithm for interval error $\alpha$ and failure probability $\delta$.

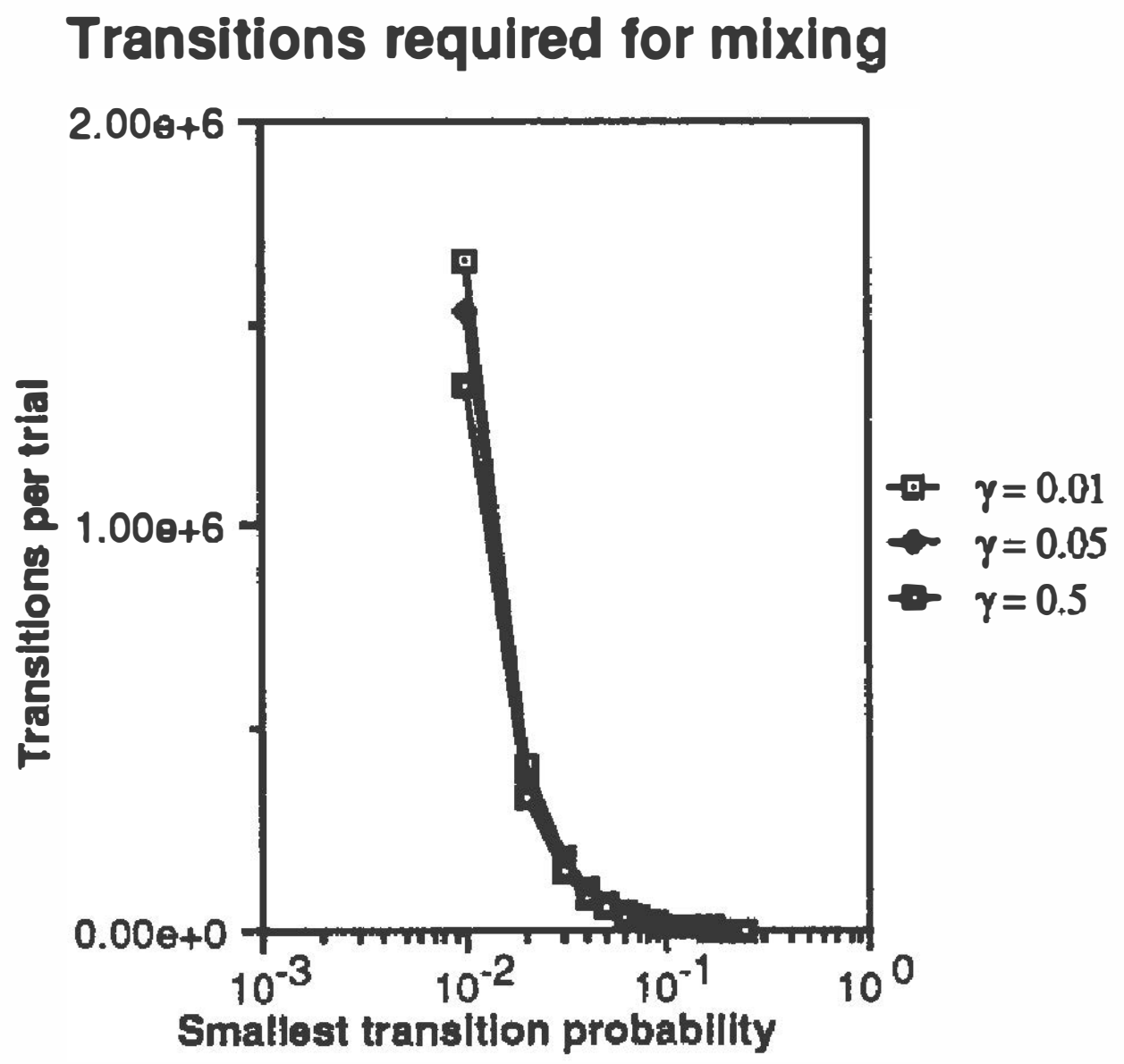

Figure 6: This graph illustrates the crucial relationship between $p_0$ and $t$, the number of transitions needed to guarantee an acceptable relative pointwise distance.

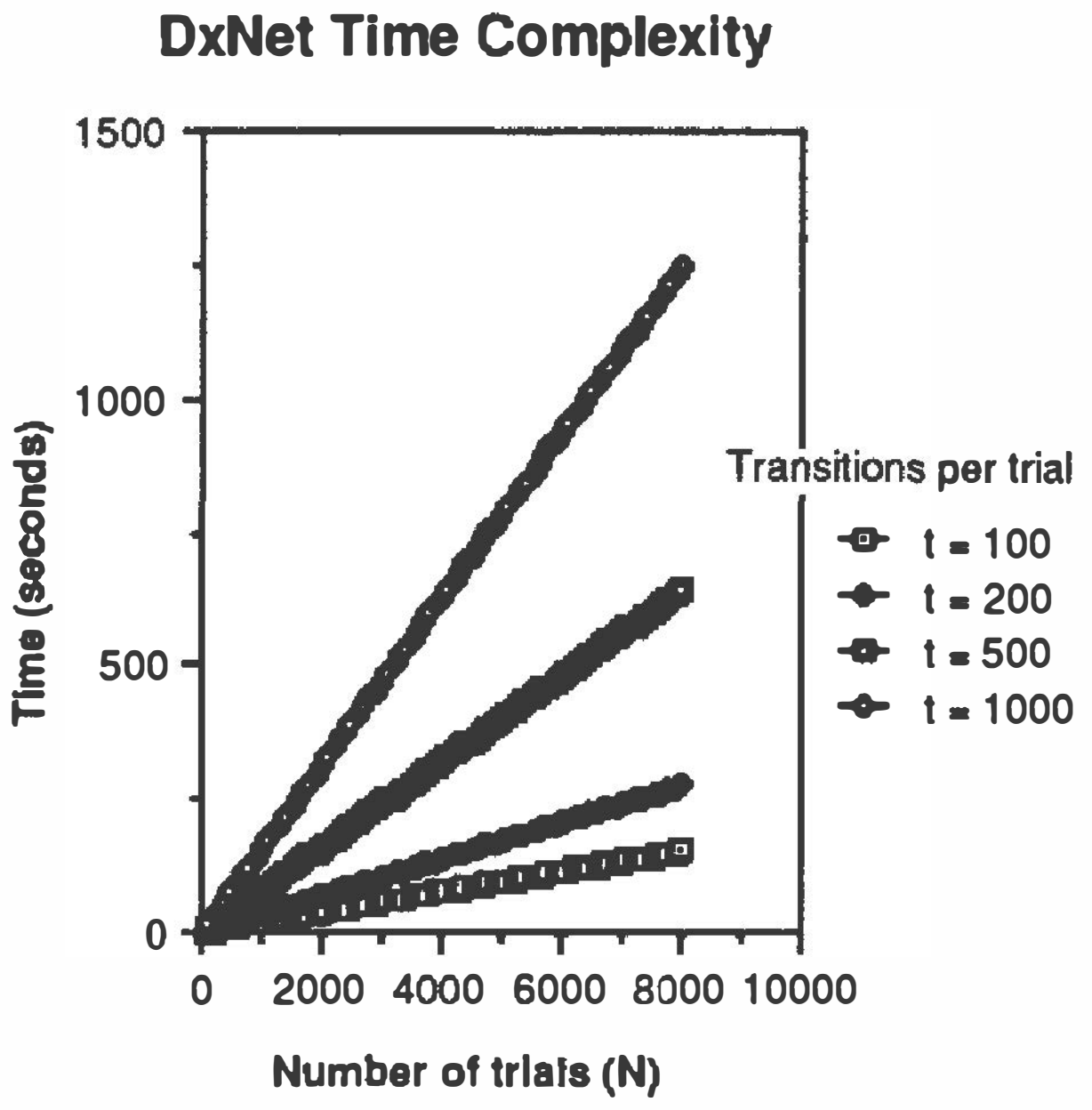

Figure 7: This graph illustrates that the computation time on a Sun-4 increases linearly with the number of trials.

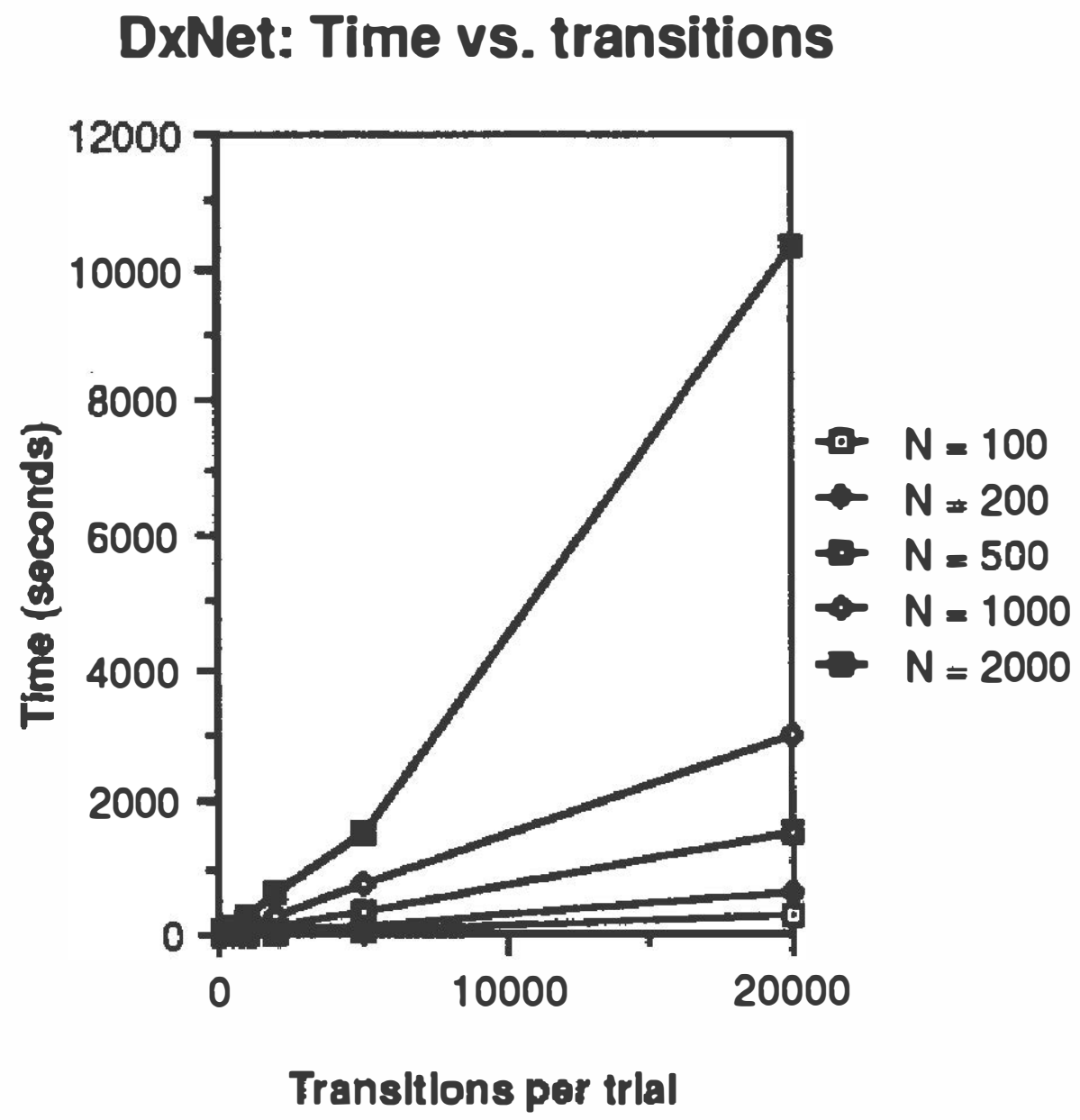

Figure 8: This graph shows the relationship between $t$, the number of transitions per trial, and the concomitant Sun-4 CPU usage.

The theoretical analysis suggests that the smallest transition probability in the Markov chain limits the rate of convergence in the worst case, as described in equation (4). For chains with large transition probabilities, we expect rapid convergence. For other networks, there is yet hope: BN-RAS and straight simulation, in contrast to the exact methods, require time linear in the number of nodes and outcomes, in the worst case. As the indegree of nodes grows, the size of cliques increases exponentially, and the Lauritzen–Spiegelhalter algorithm requires exponential time; as the loop cutset increases in size, Pearl's message-passing algorithm degrades exponentially (Suermondt and Cooper, 1988). The analysis of BN-RAS, however, indicates that the latter remains insensitive to network topology in the worst case, and degrades only as the conductance falls.

Two detailed graphs (Figures 10 and 11) make the point more cogently. Note the strong dependence of the error terms on $t$, and the absence of a close relationship between the error and $N$. These data suggest that the amount of computation required to guarantee a certain interval error depends most critically on the smallest transition probability in the network, and on very little else.

## 3.2 Comparison with Straight Simulation

BN-RAS generates $t \cdot N$ total transitions of the Markov chain, but then discards $(t-1) \cdot N$ of those states and scores only $N$ trials. In addition, the state generator shuffles the network into a random configuration at the beginning of each trial. We now compare the ras to straight simulation (Pearl, 1987b; Pearl, 1987a), both for the two-node network and for the full DxNet.

Figure 12 compares straight simulation to BN-RAS for the worst case in which the Markov chain's behavior deteriorates when the conditional probabilities approach 0 and 1. By starting with a random configuration of the network and enumerating the transitions in a fixed order, straight simulation spends most of its time looping in one state; after many transitions, the simulation falls into the other state and stays there for many transitions. Until the simulation falls into the alternative state, that state remains invisible. Hence, for networks with low conductance, straight simulations can become mired in states that serve as sinks for the Markov chain's transition probabilities (Chin and Cooper, 1987).

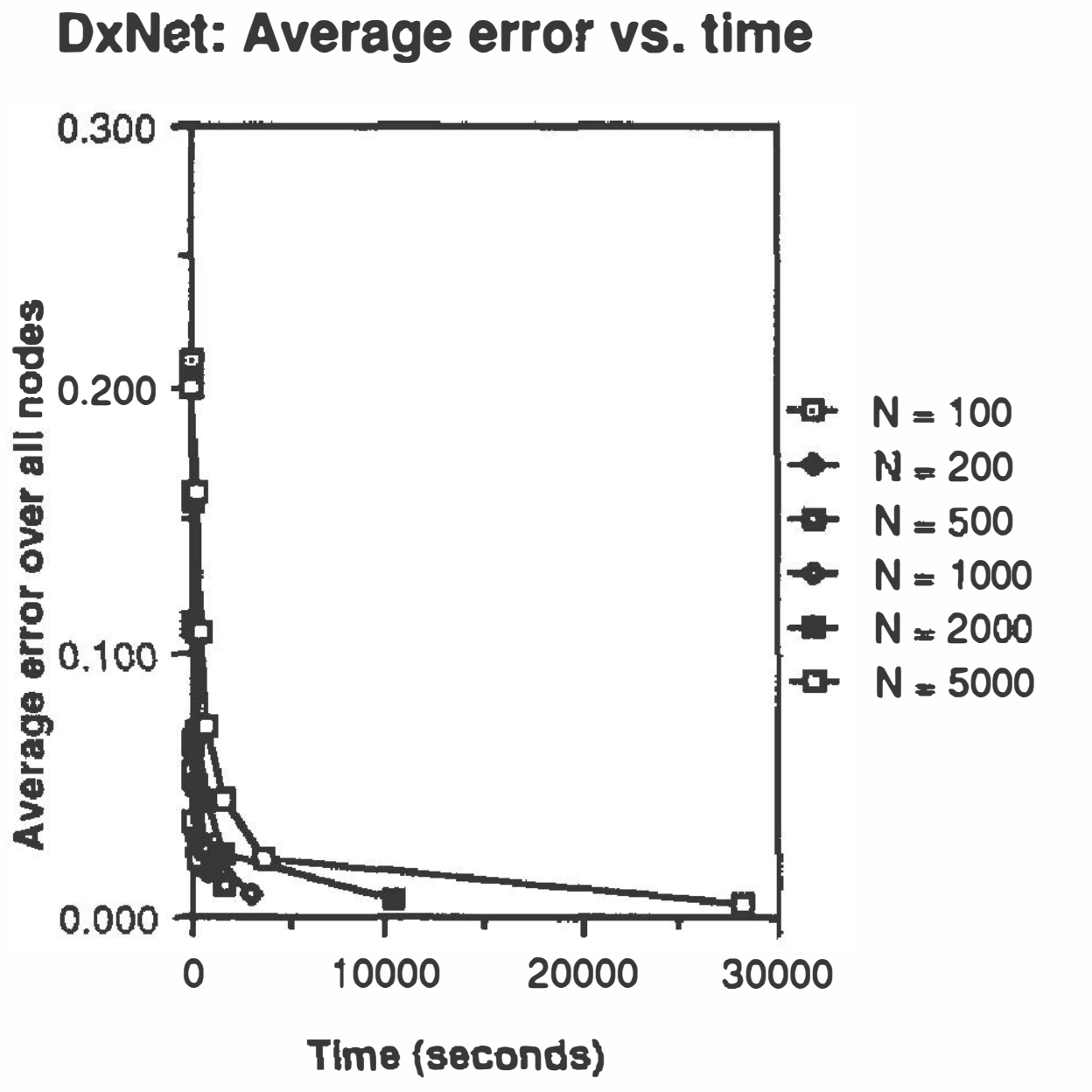

Figure 9: This graph illustrates an intriguing result: With just a few trials (on the order of 100) and many transitions per trial (on the order of 5,000 to 20,000), we can achieve rapid convergence of the average error toward 0.

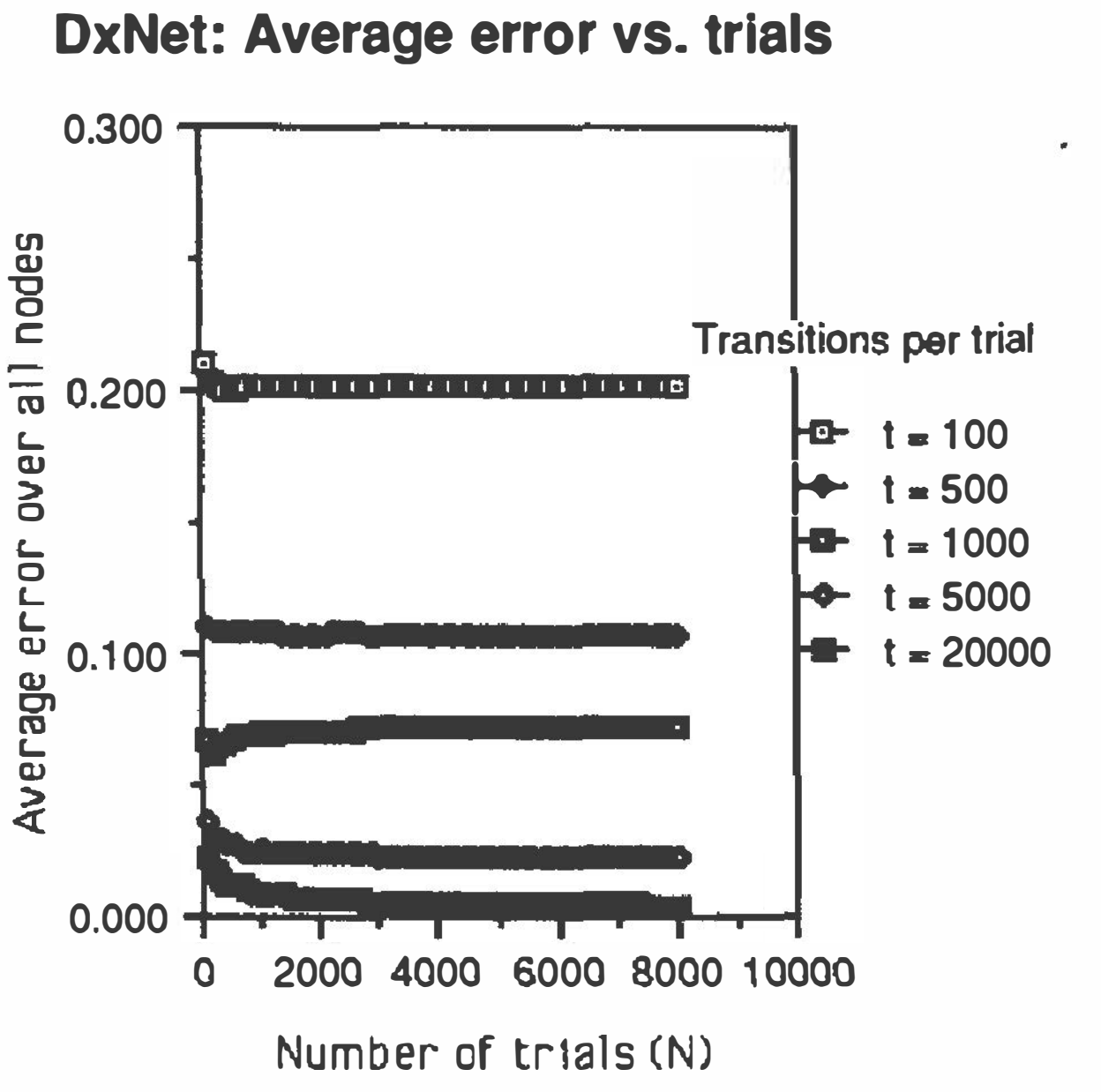

Figure 10: This graph plots the average error over all nodes against the number $N$ of trials, for different values of $t$, the number of transitions per trial. Observe that $t$ almost completely determines the convergence of the algorithm.

BN-RAS, on the other hand, randomizes the chain after $t$ trials. We therefore expect the errors to converge more uniformly toward 0, without oscillating. Indeed, Figure 12 illustrates that BN-RAS converges almost immediately to the correct answer, and stays there. We observe, however, that randomization at the beginning of each trial is not, by any means, a consistently successful strategy for improving convergence.

For the full DxNET, straight simulation and BN-RAS exhibit nearly identical convergence properties. The randomization step at the beginning of each trial in BN-RAS, and the temporally symmetric selection of transitions from an ergodic and time-reversible Markov chain, do not necessarily improve performance. Notice, however, that BN-RAS achieves the same performance as straight simulation, even though BN-RAS throws away the vast majority of its trials. Clearly, the generation of good trials (by performing many transitions) reduces error more dramatically than the scoring of many poor trials.

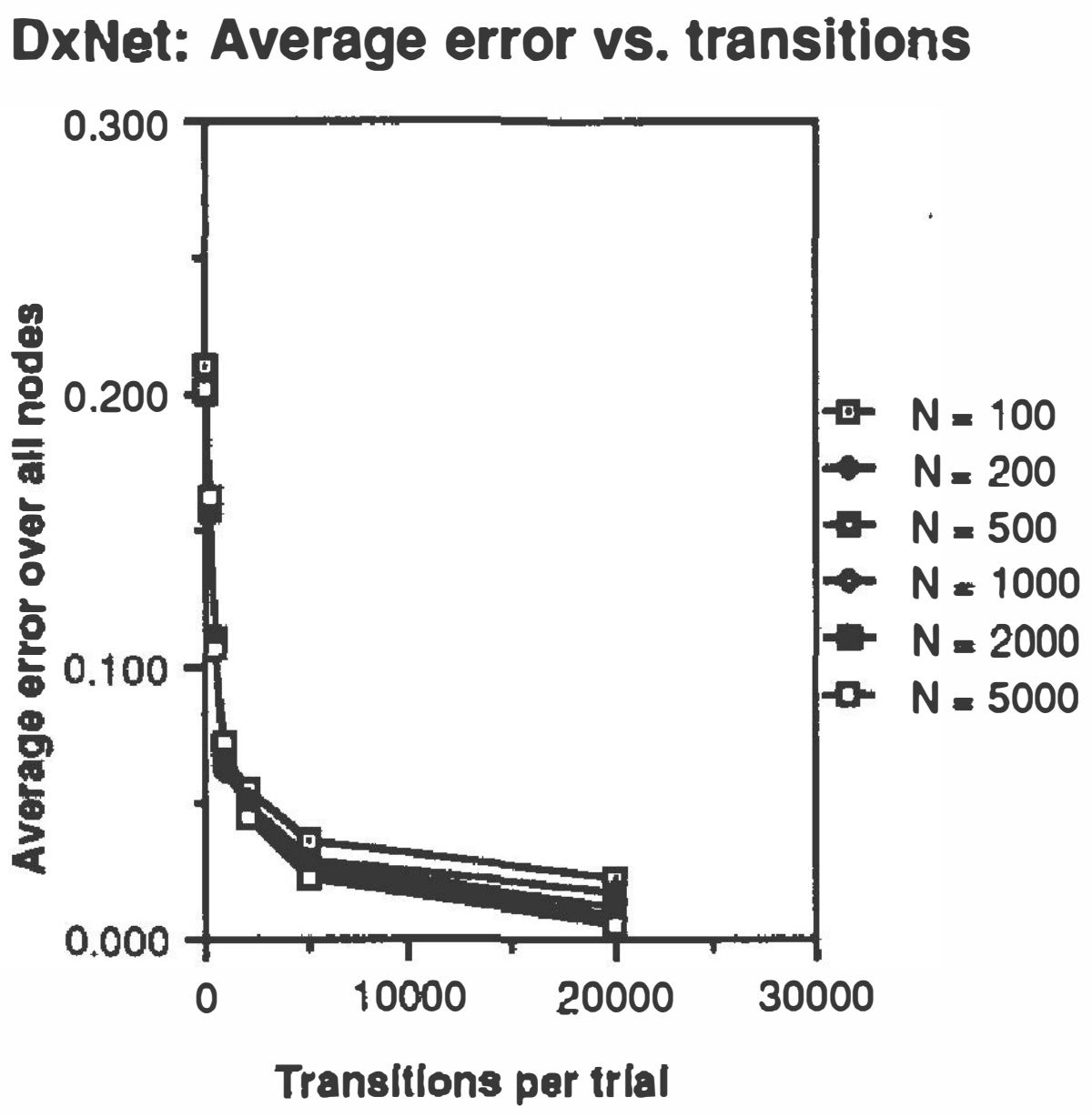

Figure 11: This graph plots the average error against the number of transitions, for different $N$, the number of trials.

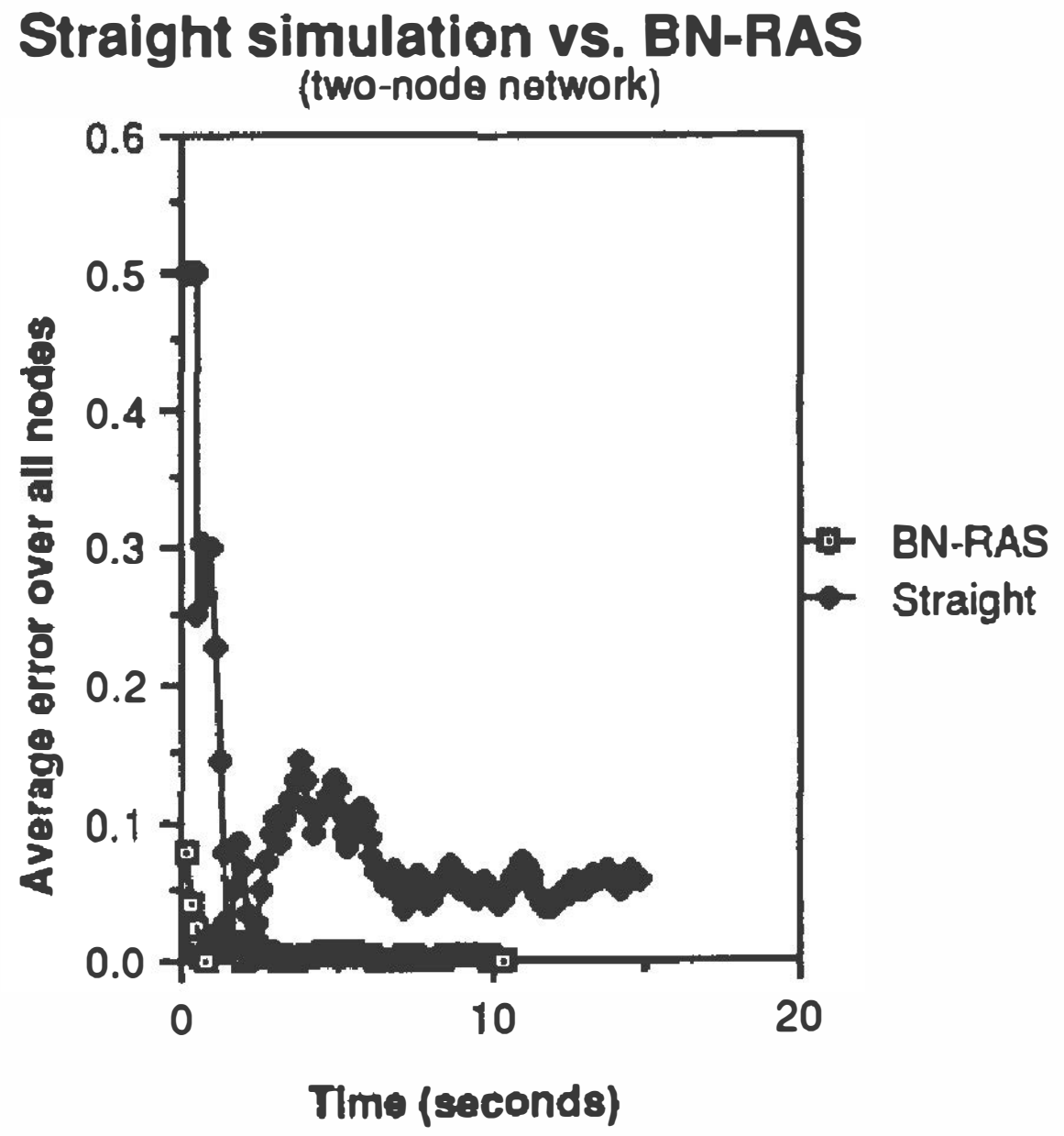

Figure 12: This graph compares the average errors for straight simulation and BN-RAS on the two-node belief network.

## 4. Discussion and Conclusions

Our investigations suggest that a precise analytic characterization of a randomized algorithm's properties can guide the search for more efficient approximations. We have shown, in particular, that the number of transitions per trial, and not the generation of a sufficient number of trials, constrains the precision of Monte Carlo approximations. Our results demonstrate that, for belief networks with transition probabilities bounded away from 0 and 1, randomized techniques offer acceptable performance. The mere generation of copious trials is not, however, likely to ensure success.

A randomized algorithm that provides a priori convergence criteria, coupled with an extensive empirical analysis, can perform efficient probabilistic inference on large networks. In addition, we have formally shown that the randomized algorithm requires time that is linear in the problem size, and polynomial in the error criteria (namely, the success probability $\delta$ and the interval error $\alpha$). As the complexity of a belief network increases, randomized algorithms may offer the only tractable approach to probabilistic inference.

We must, however, hasten to reiterate that the minimum transition probability $p_0$ severely constrains the efficacy of randomized techniques. In addition, BN-RAS does not necessarily outperform straight simulation in raw computation. In contrast to straight simulation, however, BN-RAS offers a detailed convergence analysis and a priori bounds on running time.

Finally, we outline a set of experiments in progress to characterize further the usefulness of randomized algorithms for probabilistic inference.

- We shall study the performance of the algorithm on networks of various topologies, with a particular emphasis on inference problems that cannot be solved efficiently by deterministic methods. Networks with large loop cutsets and large indegrees offer particularly severe tests of exact algorithms. We conjecture that, as long as the smallest transition probability stays the same, the ras should remain insensitive to variations in topological structure.

- We shall study the algorithm's performance on networks of different sizes and roughly similar topology (of the same maximum indegree and cutset complexity), with the smallest transition probability held constant. We expect that the performance of BN-RAS should depend very little on the size of the network. Clearly, we must expand computational resources on the order of the network's area so as to propagate an inference from one end to the other; it seems reasonable to expect, however, that the transitions per trial will dominate the running time.

## 5. Acknowledgments

This work has been supported by grant IRI-8703710 from the National Science Foundation, grant P-25514-EL from the U.S. Army Research Office, Medical Scientist Training Program grant GM07365 from the National Institutes of Health, and grant LM-07033 from the National Library of Medicine. Computer facilities were provided by the SUMEX-AIM resource under grant RR-00785 from the National Institutes of Health.

## References

I. A. Beinlich, H. J. Suermondt, R. M. Chavez, and G. F. Cooper. The ALARM monitoring system: A case study with two probabilistic inference techniques for belief networks. Technical Report KSL-88-84, Medical Computer Science Group, Knowledge Systems Laboratory, Stanford University, Stanford, CA, December 1988.

R.M. Chavez. A fully polynomial randomized approximation scheme for the Bayesian inferencing problem. Technical Report KSL-88-72, Knowledge Systems Laboratory, Stanford University, Stanford, CA, April 1989.

R.M. Chavez. *Hypermedia and randomized algorithms for probabilistic expert systems.* Ph.D. thesis proposal, Knowledge Systems Laboratory, Stanford University, Stanford, CA, January 1989. To appear in *Networks.*

H. L. Chin and G. F. Cooper. Stochastic simulation of Bayesian belief networks. In *Proceedings of the Third Workshop on Uncertainty in Artificial Intelligence*, pages 106–113, Seattle, Washington, July 1987. American Association for Artificial Intelligence.

G. F. Cooper. Probabilistic inference using belief networks is NP-hard. Technical Report KSL-87-27, Medical Computer Science Group, Knowledge Systems Laboratory, Stanford University, Stanford, CA, May 1987.

M. R. Garey and D. S. Johnson. *Computers and Intractability: A Guide to the Theory of NP-Completeness.* W. H. Freeman and Company, New York, 1979.

M. Jerrum and A. Sinclair. Conductance and the rapid mixing property for Markov chains: The

approximation of the permanent resolved. In *Proceedings of the Twentieth ACM Symposium on Theory of Computing*, pages 235–244, 1988.

R. M. Karp and M. Luby. Monte-Carlo algorithms for enumeration and reliability problems. In *Proceedings of the Twenty-fourth IEEE Symposium on Foundations of Computer Science*, 1983.

S. L. Lauritzen and D. J. Spiegelhalter. Fast manipulation of probabilities with local representations with applications to expert systems. Technical Report R-87-7, Institute of Electronic Systems, Aalborg University, Aalborg, Denmark, March 1987.

J. Pearl. Fusion, propagation, and structuring in belief networks. *Artificial Intelligence*, 29:241–288, 1986.

J. Pearl. Addendum: Evidential reasoning using stochastic simulation of causal models. *Artificial Intelligence*, 33:131, 1987.

J. Pearl. Evidential reasoning using stochastic simulation of causal models. *Artificial Intelligence*, 32:245–257, 1987.

H. J. Suermondt and G. F. Cooper. Updating probabilities in multiply connected belief networks. In *Proceedings of the Fourth Workshop on Uncertainty in Artificial Intelligence*, pages 335–343, University of Minnesota, Minneapolis, MN, August 1988. American Association for Artificial Intelligence.